\documentclass{article} 
\usepackage{iclr2016_conference,times}
\usepackage[breaklinks]{hyperref}
\usepackage{url}

\usepackage{graphicx}

\usepackage{color}
\usepackage{soul}
\usepackage{url}
\usepackage{geometry}
\usepackage{tikz}
\usepackage{tikzscale}
\usetikzlibrary{positioning}
\usetikzlibrary{shapes.misc}

\usepackage[lined,boxed,commentsnumbered]{algorithm2e}

\usepackage{booktabs}

\usepackage{epsfig}
\usepackage{float}
\usepackage{subfigure}
\usepackage[font=small]{caption}

\usepackage{amsmath,amsfonts,amssymb,amsthm}
\usepackage{mathrsfs}

\title{Alternative structures for character-level RNNs}

\author{
Piotr Bojanowski \thanks{Most of the work was done while interning at Facebook AI Research.} \\
INRIA \\
Paris, France \\
\texttt{piotr.bojanowski@inria.fr}
\AND
Armand Joulin and Tom{\'{a}}{\v{s}} Mikolov \\
Facebook AI Research \\
New York, NY, USA \\
\texttt{\{tmikolov,ajoulin\}@fb.com}
}

\newcommand{\etc}{\emph{etc.}}

\begin{document}

\maketitle


\begin{abstract}
    Recurrent neural networks are convenient and efficient models for language modeling.
    However, when applied on the level of characters instead of words, they suffer from several problems.
    In order to successfully model long-term dependencies, the hidden representation needs to be large.
    This in turn implies higher computational costs, which can become prohibitive in practice.
    We propose two alternative structural modifications to the classical RNN model.
    The first one consists on conditioning the character level representation on the previous word representation.
    The other one uses the character history to condition the output probability.
    We evaluate the performance of the two proposed modifications on challenging, multi-lingual real world data.
\end{abstract}

\section{Introduction}


Modeling sequential data is a fundamental problem in machine learning with many applications,
for example in language modeling~\citep{goodman2001bit},
speech recognition\citep{young1997htk} and machine translation~\citep{koehn2007moses}.
In particular, for modeling natural language, recurrent neural networks~(RNNs)
are now widely used and have demonstrated state-of-the-art performance in many standard tasks~\citep{mikolov12statistical}.

While RNNs have been shown to outperform the traditional n-gram models and feeedforward neural network
language models in numerous experiments, they are usually based on the word level
information and thus are oblivious to subword information.
For example, RNN language models encode input words such as ``build'', ``building'' and ``buildings''
using 1-of-N coding, which does not capture any similarity of the written form of the words.
This can potentially result in
poor representation of words that are rarely seen during training.
Even worse, the words that appear only in the test data
will be not represented at all. This problem can become significant when working with languages
that have extremely large vocabularies, such as agglutinative languages where words can
be created by concatenating morphemes (Finnish and Turkish being well-studied examples).
Further, in many real-world applications, typos and spelling mistakes artificially
increase the size of the vocabulary by adding several versions of the same word.
This requires ad-hoc spell checking approaches that are designed disjointly from
the main language modeling task.

To overcome these limitations, we investigate the use of character based recurrent neural networks (Char-RNNs) to capture subword information.
While this type of models has been widely studied in the past (see for example~\citep{mikolov2011rnnlm,sutskever2011generating,graves13generating}),
Char-RNNs lead to both lower accuracy and higher computational cost than word-based models~\citep{mikolov12subword}.
This drop of performance is unlikely due to the difficulty for character level model to capture longer short term memory,
since also the Longer Short Term Memory (LSTM) recurrent networks work better with word-based input~\citep{graves13generating}.

Ad-hoc solutions based on larger sub-word units seem to be able to both deal with new words and offer reasonable
accuracy and training speed~\citep{mikolov12subword}. However, these approaches have several issues:
one has to specify how to create the sub-word units, which can differ from language
to language; and a word can have multiple segmentations into the sub-word units.

In this paper, we investigate at first an extension of a standard Char-RNN that includes both
word level and character level information. Arguably, such approach is simpler than the one based on sub-words,
and does not have the potential problems mentioned above.
Further, we can see that one of the fundamental differences between the word level and character level models
is in the number of parameters the RNN has to access during the training and test. The smaller is the input
and output layer of RNN, the larger needs to be the fully connected hidden layer, which makes
the training of the model expensive. Following this observation, we investigate another architecture
of the Char-RNN that does not include the (still somewhat ad-hoc) word level information,
and rather attempts to make the computation the model performs more sparse. In our experiments,
this is achieved by conditioning the computation of the probability distribution in the output
layer using the recent history. This greatly increases the number of parameters in the model
without increasing the size of the hidden layer or the output layer, and thus does not increase
the computational complexity.


First, we describe the standard RNN in the context of character prediction problem in Sec.~\ref{sec:classical}, then
we propose two different structural modifications of this model.
The first modification, described in Sec.~\ref{sec:conditioning}, combines two networks, one working with characters at the input, and the other with words.
The second approach, described in Sec.~\ref{sec:ngrams}, attempts to increase capacity of the RNN model by conditioning the softmax output on the recent history. 

\subsection*{Related work}

Agglutinative languages such as Finnish or Turkish have very large vocabularies, making
word based models impractical~\citep{kurimo2006unlimited}.
Subword units such as morphemes have been used in 
statstical models for speech recognition~\citep{vergyri2004morphology,hirsimaki2006unlimited,arisoy2009turkish,creutz2007morph}. 
In particular, \citet{creutz2007morph} show that morph-based N-gram models outperform
word based ones on most of the agglutinative languages.

A mix of word and character level input for neural network language models has been investigated
by \citet{kang2011mandarin} in the context of Chinese. 
More recently, ~\citet{kim2015character} propose a model to predict words
given character level inputs, while we predict characters based on a mix of word and character
level inputs. 

Recurrent networks have been popularized for statistical language modeling by~\citet{mikolov2010recurrent}.
Since then, many authors have investigated the use of subword units in order to deal with
Out-Of-Vocabulary (OOV) words in the context of recurrent networks. Typical choice of subword units are either
characters~\citep{mikolov2011rnnlm,sutskever2011generating,graves13generating}
or syllables~\citep{mikolov12subword}.

Others have used embedding of words to deal with OOV words~\citep{bilmes2003factored,alexandrescu2006factored,luong2013better}.
\citet{luong2013better} build word embeddings by applying a recursive neural network over morpheme embeddings, while
\citet{bilmes2003factored} build their embedding by concatenating features built on previously seen words.


\section{Simple recurrent network}
\label{sec:classical}

In this section, we describe the simple RNN model 
popularized by \citet{elman1990finding}, in the context
of language modeling.
We formulate language modeling as a discrete sequence prediction problem.
That is, we want to predict the next token in a sequence given its past. 
We suppose a fixed size dictionary of $k$ words formed from $d$ different characters.
We denote by $c_t$ the one-hot encoding of $t$-th character in the sequence, 
and $w_p$ the one-hot encoding of the $p$-th word.
Our basic unit is the character.

An RNN consists of an input layer, a hidden layer and an output layer. 
Its hidden layer has a recurrent connection which allows the propagation through time of information.
More precisely, the state of the $m$ hidden units, $h_t$ is updated as a function of its previous state, $h_{t-1}$ and the current character one-hot representation $c_t$:
$$ h_t = \sigma(Ac_t + Rh_{t-1}), $$
where $\sigma(x)\mapsto 1/(1+\exp(-x))$ is the pointwise sigmoid function, $A$ is the $m \times d$ embedding matrix and $R$ the $m \times m$ recurrent matrix. 
This hidden representation is supposed to act as a memory, and should be able to convey long-term dependencies.
With sufficiently high-dimensional hidden representation, it should be \emph{a priori} possible to store the whole history.
However, using a big hidden layer implies high computational costs which are prohibitive in practice.
Using its hidden representation, the RNN compute a probability distribution $y_t$ over the next character:
$$ y_t = f(Uh_t),$$
where $U$ is a $d \times m$ matrix and $f$ is the pointwise softmax function, i.e., $[f(x)]_i = \exp(x_i) / \sum_j \exp(x_j)$. 

\paragraph{Optimization.}

In order to learn the parameters $\theta= (A,~R,~U)$ of the model, we minimize the negative log-likelihood (NLL): 
\begin{equation}
    NLL(\theta) = - \sum_{t=1}^T c_{t+1}^\top \log y_t,
\end{equation}
with a stochastic gradient descent method and backpropagation through time~\citep{rumelhart1985learning,werbos1988generalization}.
We clip the gradient in order to avoid gradient explosion. 
The details of the implementation are given in the experiment section.


Character level RNNs have been shown to perform poorly compared to word level ones~\citep{mikolov12subword}. 
In particular, they require a massive hidden layer in order to obtain results which are on par with word
level models, this makes them very expensive to compute.
In the following sections, we describe two different structural modification
of the char-RNNs in order to add capacity while reducing the overall computational cost.

\section{Conditioning on words}
\label{sec:conditioning}

\begin{figure}[t]
    \centering
    \begin{minipage}{0.5\linewidth}
    \centering
    \begin{tikzpicture}[>=latex,text height=1.5ex,on grid]
    \tikzstyle{input}   =[circle,thick,minimum size=1cm,draw=purple!80,fill=purple!20]
    \tikzstyle{output}  =[circle,thick,minimum size=1cm,draw=green!80,fill=green!20]
    \tikzstyle{hidden}  =[circle,thick,minimum size=1cm,draw=blue!80,fill=blue!20]
 
    \node (ht) [hidden]{$h_t$};
    \node (ct) [input,left = 2cm of ht]{$c_t$};
    \node (yt) [output,right = 2cm of ht]{$y_t$};
    
    \node (gp) [hidden,below = 1.5cm of ht]{$g_p$};
    \node (wp) [input,left = 2cm of gp]{$w_p$};
    \node (up) [output,right = 2cm of gp]{$v_p$};

    \path[->]   (ht) edge node [font=\scriptsize,above] {$U$} (yt);
    \path[->]   (ct) edge node [font=\scriptsize,above] {$A$} (ht);
    \path[->]   (ht) edge [loop above] node [font=\scriptsize,above] {$R$} (ht);
    \path[->]   (gp) edge [loop below] node [font=\scriptsize,below] {$R'$} (gp);
    \path[->]   (gp) edge node [font=\scriptsize,above] {$V$} (up);
    \path[->]   (wp) edge node [font=\scriptsize,above] {$B$} (gp);

    \path[->]   (gp) edge node [font=\scriptsize,right] {$Q$} (ht);

\end{tikzpicture}
    \end{minipage}
    \begin{minipage}{0.49\linewidth}
    \centering
    \begin{tikzpicture}[>=latex,text height=1.5ex,on grid]
    \tikzstyle{input}   =[circle,thick,minimum size=1cm,draw=purple!80,fill=purple!20]
    \tikzstyle{output}  =[circle,thick,minimum size=1cm,draw=green!80,fill=green!20]
    \tikzstyle{hidden}  =[circle,thick,minimum size=1cm,draw=blue!80,fill=blue!20]
    \tikzstyle{product} =[cross out,thick,minimum size=1cm,draw=black, inner sep=0pt, outer sep=0pt]
    \tikzset{
        cross/.style={
        path picture={ 
            \draw[black]
            (path picture bounding box.south east) -- (path picture bounding box.north west) (path picture bounding box.south west) -- (path picture bounding box.north east);
        }
        }
    }
    
    \node (ht) [hidden]{$h_t$};
    \node (ct) [input,left = 1.65cm of ht]{$c_t$};
    \node (prod) [draw,circle,cross,minimum size=0.7cm,right = 1.5cm of ht]{};
    \node (yt) [output,right = 1.5cm of prod]{$y_t$};
    \node (gt) [input,below = 1.5cm of prod]{$n_t$};

    \path[->]   (ht) edge node [font=\scriptsize,above] {} (prod);
    \path[->]   (gt) edge node [font=\scriptsize,above] {} (prod);
    \path[->]   (prod) edge node [font=\scriptsize,above] {$U$} (yt);
    \path[->]   (ct) edge node [font=\scriptsize,above] {$A$}  (ht);
    \path[->]   (ht) edge [loop below] node [font=\scriptsize,below] {$R$} (ht);

\end{tikzpicture}
    \end{minipage}
    \caption{
        Illustration of the two network modifications that we propose in this paper.
        \textbf{(Left)} mixed model described in Sec.~\ref{sec:conditioning}.
        The ``faster'' character-level network is conditioned on the hidden representations of a ``slower'', word-level one.
        \textbf{(Right)} conditional model described in Sec.~\ref{sec:ngrams}.
        The output at every time step $t$ depends not only on the hidden representation $h_t$ but also on the input history $n_t$.
    }
    \label{fig:tikz-words}
\end{figure}
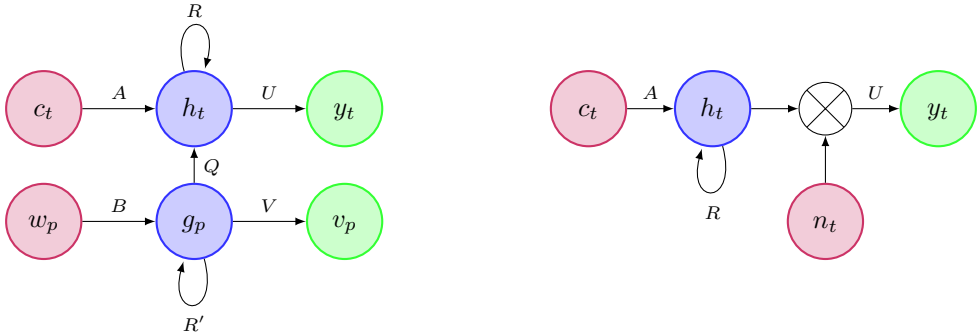

In this section, we consider an extension of character-level RNN by conditioning it with word level information.
This allows a more direct flow of information from the previous words to the character level prediction.
We propose to condition the character level on a context vector $z_t$ as follow:
\begin{equation}
    h_t = \sigma(A c_t + R h_{t-1} + Q z_t),
\end{equation}
where $Q$ is the conditioning matrix.
The context vector $z_t$ is built by gathering information at the word level using a word level RNN, with a similar architecture to the one described in the previous section.
Its input for the $p$-th word is its one-hot representation $w_p$.
The context vector is then simply the state of the hidden layer $g_p$: if the $t$-th character belongs to the $p$-th word, then $z_t = g_{p-1} $.
Figure~\ref{fig:tikz-words} \textbf{(Left)} provides an illustration of this hybrid word and character level RNN.

In order to train this model, we combine a loss on characters with a loss on words. 
However, computing a full softmax at the word level is expensive, in particular when 
facing large vocabularies.
Many solutions have been proposed to reduce the cost of this step, such as hierachical softmax, or sampling techniques.
In this work, we simply restrict the word vocabulary for the output of the word level RNN.
We keep the most frequent words (between 3000 and 5000) and associate the other ones with the $\texttt{<UNK>}$ token.
The loss that we use to train our model can be written as:
\begin{equation}
    \text{NLL}(\theta) = - \lambda \sum_{t=1}^T c_{t+1}^\top \log y_t - (1-\lambda) \sum_{p=1}^P  w_{p+1}^\top \log v_p,
\end{equation}
where $v_p$ is the prediction made by the word level RNN and $\lambda > 0$ is an interpolation parameter. 
Note that the restricted vocabulary is only used for the output of the word level RNN, the rest of the model works on the large vocabulary.
In the next section, we describe another structural modification that we propose to speed up language modeling on the character level.

\section{Conditioning prediction on recent history}
\label{sec:ngrams}

In a character based RNNs, the classifier has a very small number of parameters. We propose to condition this
classifier on the recent contextual information to increase its capacity while keeping the computational cost constant.
This context information is related to simple short-term statitics, such as coocurrence 
in a language. 
Explicitly modeling this information in the classifier, removes some of the burden from the hidden layer, allowing to use smaller
recurrent matrices while maintaining the performance.

There are many relevant contextual informations on which we can condition the classifier. 
In particular, simple short term dependencies are easily captured by n-grams, which are memory expensive but very efficient.
Using such cheap information to condition the classifier of the RNN gives a simple way to increase the capacity of the model while encouraging the rest of the RNN
to focus on more complex stastitical patterns.
Conditioning the classifier on n-grams can be written as a bilinear model. 
More precisely, denoting by $n_t$ the one-hot representation of the $t$-th n-gram, the prediction of our model is:
$$ y_t = f( n_t^\top U h_t ),$$
where $U$ is a tensor going from the product space of hidden representation and n-grams to characters.
This alternative architecture is depicted in Fig.~\ref{fig:tikz-words}~\textbf{(Right)}.

Obviously, the exponential number of n-grams makes this model impossible to learn. Worse, it is impossible to generalize
at test time to unseen n-grams. 
To avoid these problems, we restrict our set of n-grams $\mathcal{N}$ to those that contain less than $N$ symbols
and appear frequently enough in the training set.
If multiple n-grams can be used at the $t$-th character of the text, we fix $n_t$ to be the longest one appearing in $\mathcal{N}$.
This insures that each n-gram $n_t$ is associated with enough examples to learn a statistically meaningful tensor $U$.
The great thing with this solution is the fact that apart from characters that don't appear in the training set, we always have a non-trivial output model at test time.
In the worst case, this procedure will select the model corresponding to the last unigram.
The n-gram cut-off frequence allows to control the overfitting of the model.

\section{Experimental evaluation}

We evaluate the proposed models on the Penn Treebank corpus and on a subset of the Europarl dataset.
For the sake of simplicity, we compare our models to a plain RNN but all the modifications that we propose can be applied to more complex units (LSTM \etc).
We train our model using stochastic gradient descent and select hyper parameters on a validation set.
We set a constant learning rate $\gamma$ and when the validation entropy starts to increase, we divide it by a factor $\alpha$ after every epoch (values arround $\gamma=0.1$ and $\alpha=1.5$ work well in practice).
Our implementation is a single threaded CPU code which could easily be parallelized.
Code for training both models is publicly available\footnote{https://github.com/facebook/Conditional-character-based-RNN}.

We evaluate our method using entropy in bits per character.
It is defined as the empirical estimate of the cross-entropy between the target distribution and the model output in base 2.
This corresponds to the negative log likelihood that we use to train our model up to a multiplicative factor: $BPC(\theta)~=~\frac{1}{T\log(2)}~NLL(\theta)$. 

\subsection{Experiments on Penn Treebank}
\label{sec:experiments-ptb}

We first carry out experiments on the Penn Treebank corpus~\citep{marcus93building}.
This a dataset with a training set composed of 930k normalized words, yielding a total of 5017k characters.
All characters are in the ASCII format which leads to a limited size of character vocabulary $C$.
The text was normalized and the word dictionary limited to $10000$ most frequent words of the training set. 
The other words were replaced by a \verb+<UNK>+ token in the training, validation and testing sets.

\begin{table}[t]
    \centering
    \caption{
        Performance of the proposed models as compared to a classical RNN on Penn Treebank.
        For all methods we report the entropy in bits per character on the validation set and the corresponding training time for one epoch.
    }
    \begin{tabular}{rrrrrrrrr}
        \toprule
        && \multicolumn{3}{c}{val BPC} && \multicolumn{3}{c}{training time (s / epoch)} \\
        \cmidrule{3-5} \cmidrule{7-9}
        $m$ && CRNN & Mixed & Cond. && CRNN & Mixed & Cond. \\
        \midrule
         100 && 1.86 & 1.73 & 1.51 &&  166  &  613 &  250 \\ 
         200 && 1.63 & 1.50 & 1.48 &&  527  & 1152 &  707 \\ 
         300 && 1.53 & 1.43 & 1.47 && 1061  & 1838 & 1360 \\ 
         500 && 1.46 & 1.40 & 1.46 && 2645  & 3722 & 3237 \\ 
        1000 && 1.42 &  N/A &  N/A && 9752  &  N/A &  N/A \\ 
        \bottomrule
    \end{tabular}
    \label{tab:mixed-penn}
\end{table}

We evaluate both models described in this paper on the Penn Treebank dataset.
For the mixed model from Sec.~\ref{sec:conditioning} (Mixed), we fix size of the word-level hidden representations to 200. 
For the conditional model presented in Sec.~\ref{sec:ngrams} (Cond.), we choose the optimal $N$ on the validation set.
We compare these models with our own implementation of a character-level RNN as it allows us to fairly compare run times (a significant part of the code is shared).
All models are trained for various sizes of the hidden layer $m$.
We report entropy in bits per character on the validation set and the training time per epoch in Table~\ref{tab:mixed-penn}.

The character-level performance we obtain for the ``vanilla'' RNN is coherent with numbers published in the past on this dataset~\citep{mikolov12subword}.
We observe three important things: \textbf{(a)}, for any size of character hidden layer, both proposed models perform better than the plain one.
This of course comes at the expense of some additional computational cost and the benefits seem to decrease when the size of $h_t$ grows.
\textbf{(b)} using this model, we manage to obtain a comparable performance to the heavy 1000-dimensional character RNN with a hidden representation of only 300.
This corresponds to an important reduction in the number of recurent parameters and to a five times speedup per epoch at training time.
\textbf{(c)} when the hidden representation is small, the best working model is the conditional one.
However, when $m$ gets larger, the mixed one seems to work best, and provides competitive entropy for a reasonable runtime. 

For the conditional model, we observed in our experiments, that there seems to be a clear trade off in the choice of $N$, reached on Penn Treebank at roughly $N=1000$.
Indeed, in the limit case, when $N$ is very large, we only have one output model, and the network is exactly equivalent to a RNN.
On the other hand, when $N$ is small, we keep a separate model for any sequence and therefore overfits to the training set.

\subsection{Binary Penn Treebank}

We carry out some experiments on the binary representation of Penn Treebank.
As mentionned in the introduction, we would like to develop models that would be independent of the representation in use.
Working with binary representation would allow to have models of sequential data that would be agnostic to the nature of the sequence.
This could straightforwadly be applied to language modelling but also speech recognition directly from wave files \etc

We run the conditional model and a baseline bit-level RNN, both with a hidden representation of $100$.
For the conditional model we select the optimal $N$ by choosing it on the validation set ($N=2000$).
We evaluate both models by computing the entropy per bit and per character.
The results for this experiment are presented in Table~\ref{tab:binary-cond}.
This setting corresponds to the extreme case where the dictionary is as small as it could be.
The input and output model only have $2 \times m$ which can be a serious limitation for RNNs.
As we see in Table~\ref{tab:binary-cond}, the conditional model works much better as it compensates this small output model by storing several ones instead.

\begin{table}[t]
    \centering
    \caption{
        Comparison of the conditional model to a RNN for the binary representation of Penn Treebank.
        We provide results in Bits Per Bit and Bits per character.
        The conditional model outperforms the plain RNN by a large margin for a fixed size of hidden representation.
    }
    \begin{tabular}{rrrrrrr}
        \toprule
                        && \multicolumn{2}{c}{BPB} && \multicolumn{2}{c}{BPC}\\
        \cmidrule{3-4} \cmidrule{6-7}
        model && val & test && val & test  \\
        \midrule
        $\text{CRNN}$   && 0.287 & 0.282 && 2.29 & 2.25 \\
        Cond.           && 0.222 & 0.216 && 1.78 & 1.73 \\
        \bottomrule
    \end{tabular}
    \label{tab:binary-cond}
\end{table}

\subsection{The multilingual Europarl dataset}
\label{sec:europarl}

We perform another set of experiments on the Europarl dataset~\citep{koehn05europarl}.
It is a corpus for machine translation with sentences from 20 different languages aligned with their English correspondence.
For almost every language, there is more than 500k sentences, composed of more that 10M words.
Because of its size, we restrict our experiments to a subset of sentences for each language.
We randomly permute lines of the transcriptions, select 60k sentences for training, 10k for validation and 10k for testing.
The permutation we use will be made publicly available upon publication.

In this experiment, as in the one described in Sec.~\ref{sec:experiments-ptb}, we compare our models to a character-level RNN.
We train our mixed model with a word hidden of $200$ and a character hidden of $300$.
For the conditional one, we fix the hidden representation and select the optimal $N$ on the validation set.
As a baseline, we train a character-level RNN for two sizes of hidden layers: $200$ and $500$.
These results are summarized in Table~\ref{tab:mixed-europarl}, where we group ``light'' and ``heavy'' models together.
We also report the word dictionary size ($k$) and out of vocabulary rate (OOVR) for every language.

\begin{table}[t]
    \centering
    \caption{
    	Results on the Europarl dataset.
    	For all languages, we report the word vocabulary size and the out-of-vocabulary rate on the validation set.
        We report the performance of the mixed model (\textbf{Mixed}) and the conditional model (\textbf{Cond.}).
        We compare to a character-level RNN (\textbf{CRNN}) with hidden representations of size 200 and 500.
    }
    \begin{tabular}{rrrrrrrrrr}
        \toprule
        &&       &       && \multicolumn{2}{c}{large models} && \multicolumn{2}{c}{light models} \\
        \cmidrule{6-7} \cmidrule{9-10}
        language    && Vocab. size & OOVR && $\text{CRNN}_{500}$ & Mixed && $\text{CRNN}_{200}$ & Cond. \\
        \midrule
        Bulgarian 	&& 109 k & 1.87 && 1.28 & 1.26 && 1.52 & 1.27 \\
		Czech 		&& 144 k & 3.02 && 1.54 & 1.53 && 1.79 & 1.52 \\
		Danish 		&& 128 k & 2.78 && 1.37 & 1.36 && 1.62 & 1.37 \\
		German 		&& 136 k & 2.78 && 1.32 & 1.31 && 1.53 & 1.30 \\
		Greek 		&& 132 k & 2.24 && 1.28 & 1.27 && 1.55 & 1.27 \\
		Spanish 	&& 105 k & 1.71 && 1.28 & 1.26 && 1.50 & 1.27 \\
		Estonian 	&& 190 k & 5.29 && 1.48 & 1.50 && 1.72 & 1.47 \\
		Finnish 	&& 227 k & 6.91 && 1.39 & 1.43 && 1.63 & 1.38 \\
		French 		&& 105 k & 1.67 && 1.24 & 1.23 && 1.48 & 1.24 \\
		Hungarian 	&& 208 k & 5.05 && 1.39 & 1.42 && 1.65 & 1.36 \\
		Italian 	&& 115 k & 1.95 && 1.30 & 1.29 && 1.52 & 1.29 \\
		Lithuanian 	&& 163 k & 4.25 && 1.45 & 1.46 && 1.69 & 1.45 \\
		Latvian 	&& 138 k & 3.06 && 1.42 & 1.41 && 1.65 & 1.42 \\
		Dutch 		&& 109 k & 2.09 && 1.33 & 1.32 && 1.56 & 1.31 \\
		Polish 		&& 153 k & 3.13 && 1.41 & 1.39 && 1.66 & 1.39 \\
		Portuguese 	&& 110 k & 1.88 && 1.33 & 1.30 && 1.54 & 1.32 \\
		Romanian 	&& 104 k & 1.66 && 1.29 & 1.26 && 1.54 & 1.27 \\
		Slovak 		&& 145 k & 2.95 && 1.47 & 1.45 && 1.74 & 1.46 \\
		Slovene 	&& 132 k & 2.66 && 1.47 & 1.44 && 1.70 & 1.47 \\
		Swedish 	&& 130 k & 2.85 && 1.40 & 1.39 && 1.64 & 1.38 \\
        \midrule
        Average     &&        &      && 1.37 & 1.36 && 1.61 & 1.36 \\
        \bottomrule
    \end{tabular}
    \label{tab:mixed-europarl}
\end{table}

The CRNN baseline as well as the proposed models still are quite far from the performance of a word-level RNN.
As we see in Table~\ref{tab:mixed-europarl}, the average performance of both proposed models give a per-character entropy of 1.36.
The proposed structural modifications allow us to achieve similar performance to a large character-level RNN with a reduced computational cost.
For languages such as Finnish and Hungarian, the conditional model (\textbf{Cond.}) yields best performance.

For reference, we computed a word-level RNN baseline using a modified version of SCRNN\footnote{https://github.com/facebook/SCRNNs}.
If we assign 4 times the average entropy to OOV words, it gives us an entropy of 1.27 BPC.
The proposed models allow us to efficiently tackle the problem of learning small vocabulary sequences.
However, the gap between the word and character-level models is far from being closed.


\section*{Conclusion}

In this work we investigated modifications of RNNs for general discrete sequence prediction when the number of symbols is very small, such as in char-RNNLM. 
We found that with certain tricks, one can train the model much faster, and overall we observed that the fully connected RNN architecture has its weaknesses, especially related to the excessive computational complexity. 
We believe more research is needed to develop general mechanisms that would allow us to train RNNs with richer internal structure. 
We expect such research can greatly simplify many pipelines, for example in the NLP applications where we could avoid having separate systems that perform spell checking, text normalization, and modeling of the language disjointly.

We hope that this work will open up new research paths for modeling sequences with small vocabularies.
Initial yet promising results on binary representations of Penn Treebank show that RNNs can be trained on that kind of data.
We believe that this allows us to define models for sequence modeling that would be agnostic to the nature of the input.
Bit-level models could be used on any sequential data, for example speech signal in binary .wav form.

%

\bibliography{iclr2016_conference}
\bibliographystyle{iclr2016_conference}

\end{document}